\documentclass[runningheads]{llncs}
\usepackage{amsmath,amssymb} 
\usepackage{graphicx}

\mathchardef\mhyphen="2D
\usepackage{tabularx}
\newcolumntype{K}[1]{>{\centering\arraybackslash}p{#1}}

\begin{document}

\title{Dynamic Temporal Pyramid Network: \\A Closer Look at Multi-Scale Modeling for Activity Detection} 
\titlerunning{Dynamic Temporal Pyramid Network} 

\author{Da Zhang\inst{1} \and
Xiyang Dai\inst{2} \and
Yuan-Fang Wang\inst{1}}

\authorrunning{D. Zhang et al.}
\institute{University of California, Santa Barbara \and
University of Maryland, College Park \\
\email{dazhang@cs.ucsb.edu, xdai@umiacs.umd.edu, yfwang@cs.ucsb.edu}}

\maketitle

\begin{abstract}
Recognizing instances at varying scales simultaneously is a fundamental challenge in visual detection problems. While spatial multi-scale modeling has been well studied in object detection, how to effectively apply a multi-scale architecture to temporal models for activity detection is still under-explored. In this paper, we identify three unique challenges that need to be specifically handled for temporal activity detection. To address all these issues, we propose Dynamic Temporal Pyramid Network (DTPN), a new activity detection framework with a multi-scale pyramidal architecture featuring three novel designs: (1) We sample frame sequence dynamically with different frame per seconds (FPS) to construct a natural pyramidal representation for arbitrary-length input videos. (2) We design a two-branch multi-scale temporal feature hierarchy to deal with the inherent temporal scale variation of activity instances. (3) We further exploit the temporal context of activities by appropriately fusing multi-scale feature maps, and demonstrate that both local and global temporal contexts are important. By combining all these components into a uniform network, we end up with a single-shot activity detector involving single-pass inferencing and end-to-end training. Extensive experiments show that the proposed DTPN achieves state-of-the-art performance on the challenging ActvityNet dataset. 
\end{abstract}

\section{Introduction}
\label{sec:intro}
Temporal activity detection has drawn increasing interests in both academic and industry communities due to its vast potential applications in security surveillance, behavior analytics, videography and so on. Different from activity recognition, which only aims at {\em classifying} the categories of {\em manually trimmed} video clips, activity detection is {\em localizing} and {\em recognizing} activity instances from {\em long, untrimmed} video streams. This makes the task substantially more interesting and challenging. With recent advances in deep learning, there has been fruitful progress in video analysis. While the performance of activity recognition has improved a lot~\cite{tran2015learning,tran2018closer,simonyan2014two,carreira2017quo,qiu2017learning,dai2018tan}, the detection performance still remains unsatisfactory~\cite{shou2017cdc,dai2017temporal,lin2017single,chao2018rethinking,zhang2018s3d}. 

One major obstacle that people are facing in temporal activity detection, is how to effectively model activities with various temporal length and frequency. Especially, the challenge of localizing precise temporal boundaries among activities of varying scales has been demonstrated as one major factor behind the difference in performance \cite{dai2017temporal}. Luckily, the problem of scale variation is not new in computer vision researches, as it has been well studied in object detection in images~\cite{singh2018analysis}. In order to alleviate the problems arising from scale variation and successfully detect objects at multiple scales, extensive analysis has been conducted in recent years. Multi-scale pyramidal architecture has been widely adopted and become a general structure in many state-of-the-art object detection frameworks~\cite{liu2016ssd,lin2017feature}.

How to effectively model the temporal structure for activity detection using a multi-scale pyramidal network then? To answer this question, we first identify three unique problems that need to be specifically handled for temporal activity detection: (1) The duration of the input video is arbitrary (usually ranges from few seconds to few minutes). A naive subsampling method (resize the video) or sliding window (crop the video) will fail to fully exploit the temporal relations. (2) The temporal extent of activities varies dramatically compared to the size of objects in an image, posing a challenge to deal with large instance scale variation. (3) The spatial context of a bounding box is important to correctly classify and localize an object, and the temporal context is arguably more so than the spatial context. Thus, cross-scale analysis becomes much more crucial in temporal domain. In this work, we propose a multi-scale pyramidal deep-learning architecture with three novel elements designed to solve the above problems accordingly. 
\begin{enumerate}
  \item \textbf{How to effectively extract a feature representation for input video of arbitrary length?} A common practice in most existing works~\cite{xu2017r,chao2018rethinking,dai2017temporal} is to use a high-quality video classification network for extracting a feature representation from raw frame sequence. However, when dealing with input video of arbitrary length, they only decode the video at a fixed FPS and extract features with a single resolution. To fully exploit temporal relations at multiple scales and effectively construct a feature representation, we propose to use dynamic sampling to decode the video at varying frame rates and construct a pyramidal feature representation. Thus, we are able to parse an input video of arbitrary length into a fixed-size feature pyramid without losing short-range and long-range temporal structures. Nevertheless, our extraction method is very general and can be applied to any framework and compatible with a wide range of network architectures. 
 \item \textbf{How to build better temporal modeling architectures for activity detection?} In dealing with the large instance scale variation, we draw inspirations from SSD~\cite{liu2016ssd} to build a multi-scale feature hierarchy allowing predictions at different scales by appropriately assigning default spans. This multi-scale architecture enforces the alignment between the temporal scope of the feature and the duration of the default span. Besides, we also draw inspirations from Faster-RCNN~\cite{ren2015faster} to use separate features for classification and localization since features for localization should be sensitive to pose variation while those for classification should not. We propose a new architecture to leverage the efficiency and accuracy from both frameworks while still maintaining a single shot design. In our work, we use separate temporal convolution and temporal pooling branches with matched temporal dimension at each scale, and use a late fusion scheme for final prediction.
  \item \textbf{How to utilize local and global temporal contexts?} We claim both local temporal context (\emph{i.e.}, moments immediately preceding and following an activity) and global temporal context (\emph{i.e.}, what happens during the whole video duration) are crucial. We propose to explicitly encode local and global temporal contexts by fusing features at appropriate scales in the feature hierarchy. 
\end{enumerate}
\noindent 
Our contributions are: (1) We take a closer look at multi-scale modeling for temporal activity detection and identify three unique challenges. (2) To address all these issues in a single network, we introduce the Dynamic Temporal Pyramid Network (DTPN), which is a single shot activity detector featuring a novel multi-scale pyramidal architecture design. (3) Our DTPN achieves state-of-the-art performance on temporal activity detection task on ActivityNet benchmark~\cite{caba2015activitynet}. 

\section{Related Work}
Here, we review relevant works in activity recognition, multi-scale pyramidal modeling, and temporal activity detection. 

\noindent \textbf{Activity Recognition.} 
Activity recognition is an important research topic and has been extensively studied for a long time. In the past few years, tremendous progress has been made due to the introduction of large datasets~\cite{caba2015activitynet,jiang2014thumos} and the developments on deep neural networks~\cite{tran2015learning,tran2018closer,simonyan2014two,carreira2017quo,qiu2017learning}. Two-stream network~\cite{simonyan2014two} learned both spatial and temporal features by operating 2D ConvNet on single frames and stacked optical flows. C3D~\cite{tran2015learning} used Conv3D filters to capture both spatial and temporal information directly from raw video frames. More recently, improvements on top of the C3D architecture~\cite{tran2018closer,carreira2017quo,qiu2017learning} as well as advanced temporal building blocks such as non-local modules~\cite{NonLocal2018} were proposed to further boost the performance. However, the assumption of well-trimmed videos limits the application of these approaches in real scenarios, where the videos are usually long and untrimmed. Although they do not consider the difficult task of localizing activity instances, these methods are widely used as the backbone network for the detection task.

\noindent \textbf{Multi-scale Pyramidal Modeling.}
Recognizing objects at vastly different scales is a fundamental challenge in computer vision. To alleviate the problems arising from scale variation, multi-scale pyramidal modeling forms the basis of a standard solution~\cite{adelson1984pyramid} and has been extensively studied in the spatial domain. For example, independent predictions at layers of different resolutions are used to capture objects of different sizes~\cite{cai2016unified}, training is performed over multiple scales~\cite{he2016deep}, inference is performed on multiple scales of an image pyramid~\cite{dai2017deformable}, feature pyramid is directly constructed from the input image~\cite{lin2017feature}.  

Meanwhile, the multi-scale modeling for temporal activity detection is still under-explored: Shou~\emph{et al.}~\cite{shou2016temporal} used a multi-scale sliding window to generate snippets of different lengths, however, such method is often inefficient during runtime due to the nature of sliding window; Zhao~\emph{et al.}~\cite{zhao2017temporal} used temporal pyramid pooling for modeling multi-scale structures without considering complex motion dynamics, since those features were directly pooled at different levels. In this paper, we provide a comprehensive study on temporal multi-scale modeling and propose an efficient end-to-end solution.

\noindent \textbf{Temporal Activity Detection.} 
Unlike activity recognition, the detection task focuses on learning how to localize and detect activity instances in untrimmed videos. The problem has recently received significant research attention due to its potential application in video data analysis. 

Previous works on activity detection mainly use sliding windows as candidates and classify video clips inside the window with activity classifiers trained on multiple features~\cite{mettes2015bag}. Many recent works adopt a proposal-plus-classification framework~\cite{buch2017sst,shou2017cdc,shou2016temporal,zhao2017temporal,gao2017cascaded} by generating segment proposals and classifying activity categories for each proposal: some of them focus on designing better proposal schemes~\cite{buch2017sst,zhao2017temporal,gao2017cascaded}, while others focus on building more accurate activity classifiers~\cite{shou2017cdc,shou2016temporal,zhao2017temporal}. Along this line of attack, Xu~\emph{et al.}~\cite{xu2017r} proposed an end-to-end trainable activity detector based on Faster-RCNN~\cite{ren2015faster}. Buch~\emph{et al.}~\cite{buch2017end} investigated the use of gated recurrent memory module in a single-stream temporal detection framework. However, all these methods rely on feature maps with a fixed temporal resolution and fail to utilize a multi-scale pyramidal architecture for handling instances with varying temporal scales.

A few very recent works~\cite{chao2018rethinking,lin2017single,dai2017temporal,zhang2018s3d} have started to model temporal scales with a multi-tower network~\cite{chao2018rethinking} or a multi-scale feature hierarchy~\cite{lin2017single,zhang2018s3d}, and incorporated temporal contextual information~\cite{dai2017temporal,chao2018rethinking}. Our method differs from all these approaches in that we identify three unique modeling problems specific to temporal activity detection and propose to solve them in one single multi-scale pyramidal network. We detail our contributions below.

\begin{figure*}[t!]
\begin{center}
\includegraphics[width=1.0\linewidth]{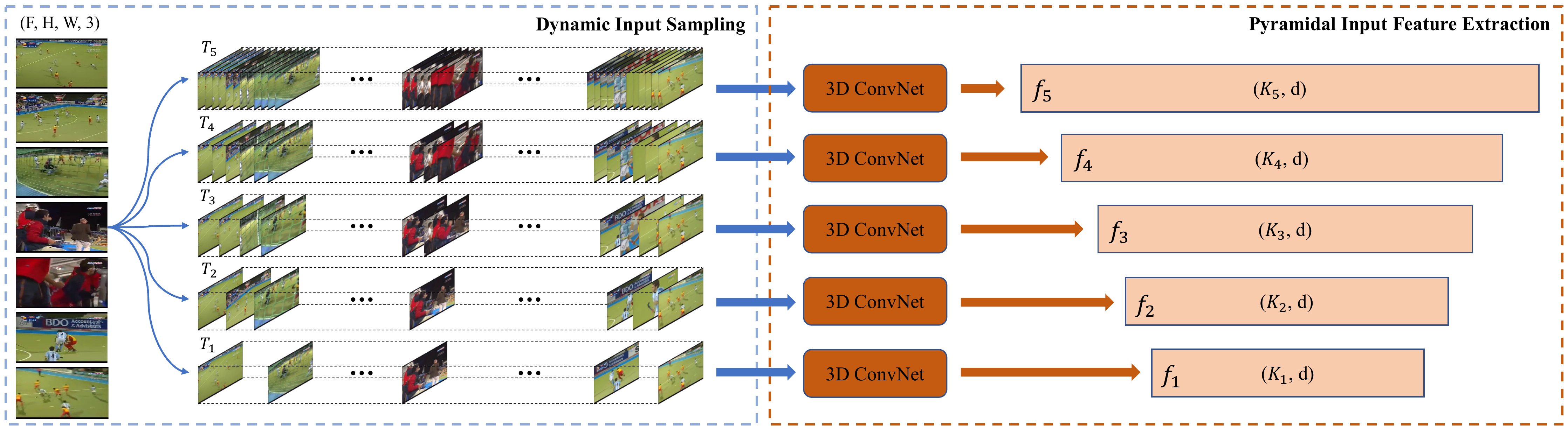}
\end{center}
   \caption{An illustration of pyramidal input feature extraction with 5 sampling rates. Left: input video is sampled at different FPS to capture motion dynamics at different temporal resolutions; Right: a shared 3D ConvNet is used to extract the input feature at each resolution.}
\label{fig:dynamic}
\end{figure*}

\section{Approach}

We present a \textit{Dynamic Temporal Pyramid Network (DTPN)}, a novel approach for temporal activity detection in long untrimmed videos. DTPN is dedicatedly designed to address the temporal modeling challenges as discussed in the introduction with a multi-scale pyramidal architecture. The overall DTPN framework is a single-shot, end-to-end activity detector featuring three novel architectural designs: pyramidal input feature extraction with dynamic sampling, multi-scale feature hierarchy with two-branch network, and local and global temporal contexts (Sec.~\ref{subsec:input} to~\ref{subsec:context}). 

\subsection{Pyramidal Input Feature Extraction with Dynamic Sampling}
\label{subsec:input}
The input of our network is an untrimmed video with an arbitrary length. We denote a video $\nu$ as a series of RGB frames $\nu=\{I_{i}\}_{i=1}^{F}$, where $I_i\in \mathbb{R}^{H\times W \times 3}$ is the $i\mhyphen th$ input frame and $F$ is the total number of frames. A common practice is to use a high-quality video classification network to extract a 1D feature representation on top of the input frame sequence~\cite{xu2017r,chao2018rethinking,lin2017single}. This feature extraction step is beneficial for summarizing spatial-temporal patterns from raw videos into high-level semantics. The backbone classification network can be of any typical architectures, including the two-stream network~\cite{simonyan2014two}, C3D~\cite{tran2015learning}, I3D~\cite{carreira2017quo}, Res3D~\cite{tran2018closer}, P3D~\cite{qiu2017learning}, etc. However, an obvious problem of the classification ConvNet in their current form is their inability in modeling long-range temporal structure. This is mainly due to their limited temporal receptive field as they are designed to operate only on a single stack of frames in a short snippet.

To tackle this issue, we propose to extract pyramidal input feature with dynamic sampling, a video-level framework to model multi-level dynamics throughout the whole video. Sparse sampling has already been proven very successful when solving the video classification problem~\cite{wang2016temporal}, where preliminary prediction results from short snippets sparsely sampled from the video are aggregated to generate the video-level prediction. Following similar ideas, we propose a general feature extraction framework specifically for the temporal activity detection.

Formally, given an input video $\nu$ with $F$ frames and a sampling scale index $s$, we divide the entire frame sequence into $K_s$ different segments of equal duration. Suppose a classification network takes $w$ frames as input and generates a $d\mhyphen dimensional$ 1D feature vector before any classification layers, we uniformly sample $w$ frames in each segment to construct a sequence of snippets $\{T_1, T_2, ..., T_{K_s}\}$, where each $T_i, i\in[1,K_s]$ is a snippet of $w$ frames which can be directly used as an input to the backbone network. Thus, we can extract features for a specific sampling scale index $s$ as 
\begin{equation}
	f_s = \bigcup_{i=1}^{K_s}F(T_i, \textbf{W}) \in \mathbb{R}^{K_s \times d}
    \label{eq:feat}
\end{equation}
where $F(T_i, \textbf{W})$ is the function representing a ConvNet with parameter $\textbf{W}$ which operates on snippet $T_i$ and generates a $d\mhyphen dimensional$ feature vector. Thus, each single feature vector $F(T_i, \textbf{W})$ in $f_s$ covers a temporal span of $\frac{F}{K_s}$ frames. Suppose the input frame sequence is decoded at $r$ FPS, then the equivalent feature-level sampling rate is given as $\frac{r\times K_s}{F}$. Instead of only extracting features at a single scale, we apply a set of different scales to construct a pyramidal input feature, which can be considered as sampling the input frame sequence with dynamic FPS. Technically, we use $S$ different scales to sample the input video with a base scale length $K_1$ and an up sampling factor of $2$. \emph{i.e.} $K_s=2^{s-1}\times K_1,s\in[1,S]$ different feature vectors will be extracted given a scale index $s$. This dynamic sampling procedure allows us to directly summarize both short-range and long-range temporal relations while being efficient during runtime. Finally, a pyramidal feature is constructed as 
\begin{equation}
	f_{pymd} = \bigcup_{s=1}^{S}f_s, f_s\in \mathbb{R}^{K_s \times d}
    \label{eq:pyramid}
\end{equation}
which will be used as the input to the two-branch network (Sec.~\ref{subsec:fusion}).

The overall procedure is illustrated in Fig.~\ref{fig:dynamic}. Note that our approach is different from temporal pyramid pooling~\cite{zhao2017temporal} where higher-level features are directly pooled, and multi-scale sliding window~\cite{shou2016temporal} where a window size is pre-defined. Our dynamic sampling approach fixes the number of sampling windows and computes independent features by directly looking at input frames with different receptive fields. We find that both sparse and dense sampling are important for temporal detection task: sparse sampling is able to model long-range temporal relations. Dense sampling, on the other hand, provides high-resolution short-range temporal features. By using an off-the-shelf video classification network and a dynamic frame sampling strategy, we are able to construct a pyramidal input feature that naturally encodes the video at varying temporal resolutions. 

\noindent \textbf{Comparison with previous works.} When extracting features from the input video, previous works~\cite{lin2017single,dai2017temporal,chao2018rethinking,xu2017r,zhang2018s3d} decode the input video with a fixed FPS (usually small for computational efficiency) and extract features using a non-overlapping sliding window, which corresponds to a fixed FPS single-scale sampling in our schema. Although advanced networks are applied to model temporal relationships, their feature extraction component fails to fully exploit the multi-scale motion context in an input video stream. More importantly, our extraction strategy is very general thus can be applied to any framework and compatible with a wide range of network architectures.

\subsection{Multi-scale Feature Hierarchy with Two-branch Network}
\label{subsec:fusion}
To allow the model to predict variable scale temporal spans, we follow the design of SSD to build a multi-scale feature hierarchy consisting of feature maps at several scales with a scaling step of $2$. We then assign default temporal spans at each layer to get temporal predictions at multiple scales. More specifically, a multi-scale feature hierarchy is created which we denote as $\{C_i\}_{i=1}^{N}, C_i\in \mathbb{R}^{{L_i}\times d_f}$ where $N$ is the total number of features each with a temporal dimension $L_i$ and feature dimension $d_f$. For a simple and efficient design, we set $L_1=K_1$ and $L_N=1$, and the temporal dimension in between follows $L_i = 2L_{i+1}$.

\begin{figure*} [t!]
\begin{center}
\includegraphics[width=1.0\linewidth]{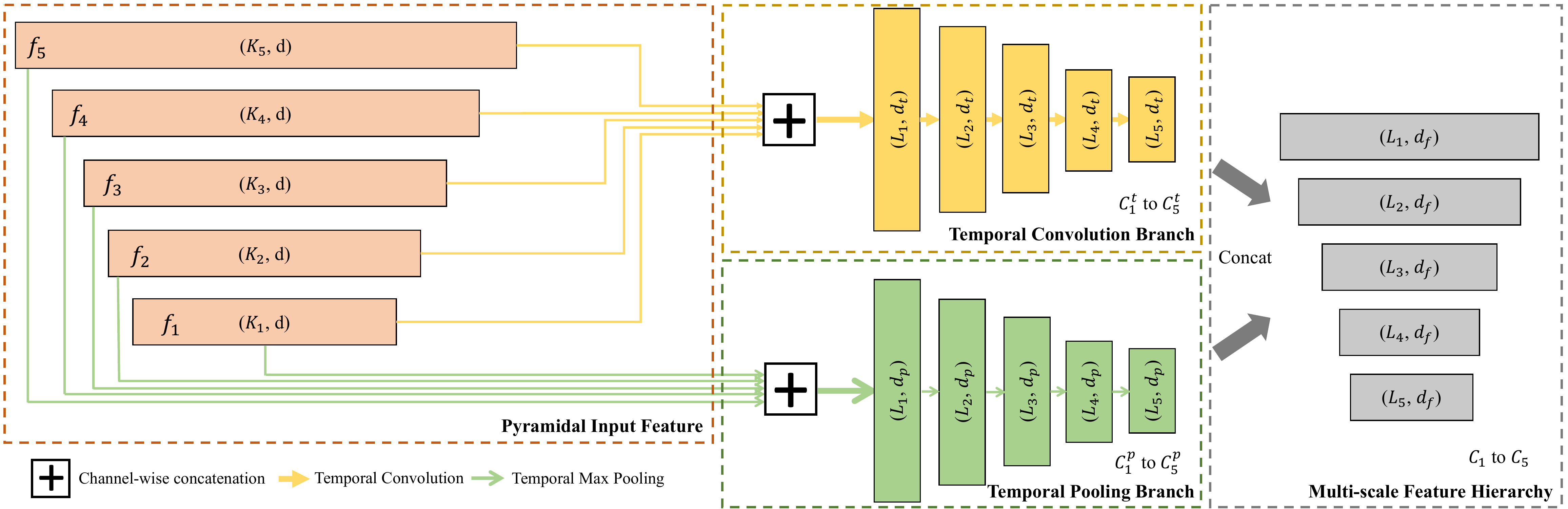}
\end{center}
   \caption{An illustration of the two-branch multi-scale network with $S=N=5$. The network combines a temporal convolution branch and a temporal pooling branch, where the features are concatenated and down sampled. Late fusion scheme is applied to build the multi-scale feature hierarchy.}
\label{fig:fusion}
\end{figure*}

The next question is: how do we combine the pyramidal input feature and build the multi-scale network? As illustrated in Fig.~\ref{fig:fusion}, we propose to use a two-branch network, \emph{i.e.}, a temporal convolution branch and a temporal pooling branch to fuse the pyramidal input feature and aggregate these branches at the end. This design choice is inspired by the fact that pooling features contain more translation-invariant semantic information which is classification-friendly and convolutional features better model temporal dynamics which are helpful for localization~\cite{ren2015faster,lin2017feature}. 

In more detail, both branches take as input the pyramidal feature $f_{pymd}$. For the temporal convolution branch, a Conv1D layer with temporal kernel size $\frac{K_s}{L_1} + 1$, stride $\frac{K_s}{L_1}$ is applied to each input feature $f_s\in f_{pymd}, s\in [1,S]$ to increase the temporal receptive field and decrease the temporal dimension to $L_1$ (temporal stride is set to $1$ for $f_1$ since no down sampling is needed). We use channel-wise concatenation to combine the resulting features into a single feature map $C^{t}_{1}\in \mathbb{R}^{L_1\times d_t}$. Based on $C^{t}_{1}$, we stack Conv1D layers with kernel size $3$ and stride $2$ for progressively decreasing the temporal dimension by a factor 2 to construct $C^{t}_{2}$ through $C^{t}_{N}$. For the temporal pooling branch, a non-overlapping temporal max pooling with window size $\frac{K_s}{L_1}$ is used on top of each input feature $f_s\in f_{pymd}, s\in [1,S]$ to match with the temporal dimension $L_1$. Similar to the temporal convolution branch, channel-wise concatenation is applied here to construct $C^{p}_{1}\in \mathbb{R}^{L_1\times d_p}$. Then, we use temporal max pooling with a scaling step of $2$ to construct the feature hierarchy $\{C_i^p\}_{i=1}^{N}$. Finally, features from the two branches are aggregated together to generate the final feature hierarchy $\{C_i\}_{i=1}^{N}$, which will be used to further model the temporal context (Sec.~\ref{subsec:context}).

Simplicity is central to our design and we have found that our model is robust to many design choices. We have experimented with other feature fusion blocks such as element-wise product, average pooling, etc., and more enhanced building blocks such as dilated convolution~\cite{yu2015multi} and observed marginally better results. Designing better network blocks is not the focus of this paper, so we opt for the simple design described above. 

\noindent \textbf{Comparison with previous works.} Previous works based on SSD framework~\cite{lin2017single,zhang2018s3d} only use a single convolutional branch and don't apply feature fusion since only a single-scale input is applied. Our design uses two separate branches with slightly different feature designs at multiple scales. The localization branch uses temporal convolution for better localization while the classification branch uses maximum pooling to record the most prominent features for recognition. We show experimentally that our two-branch design achieves much better results compared to single-branch (Sec.~\ref{subsec:ablation}). 

\subsection{Local and Global Temporal Contexts}
\label{subsec:context}
Context is important in computer vision tasks~\cite{dai2017fason,dai2017efficient}. Temporal contextual information has been shown to be critical for temporal activity detection~\cite{dai2017temporal,chao2018rethinking}. There are mainly two reasons: First, it enables more precise localization of temporal boundaries. Second, it provides strong semantic cues for identifying the activity class. In order to fully utilize the temporal contextual information, we propose to use both local temporal context (\emph{i.e.}, what happens immediately before and after an activity instance) and global temporal context (\emph{i.e.}, what happens during the whole video duration). Both contexts help with localization and classification subtasks but with different focuses: local context focuses more on localization with immediate cues to guide temporal regression, while global context tends to look much wider at the whole video to provide classification guidance. Below, we detail our approach.

\begin{figure*}[t!]
\begin{center}
\includegraphics[width=1.0\linewidth]{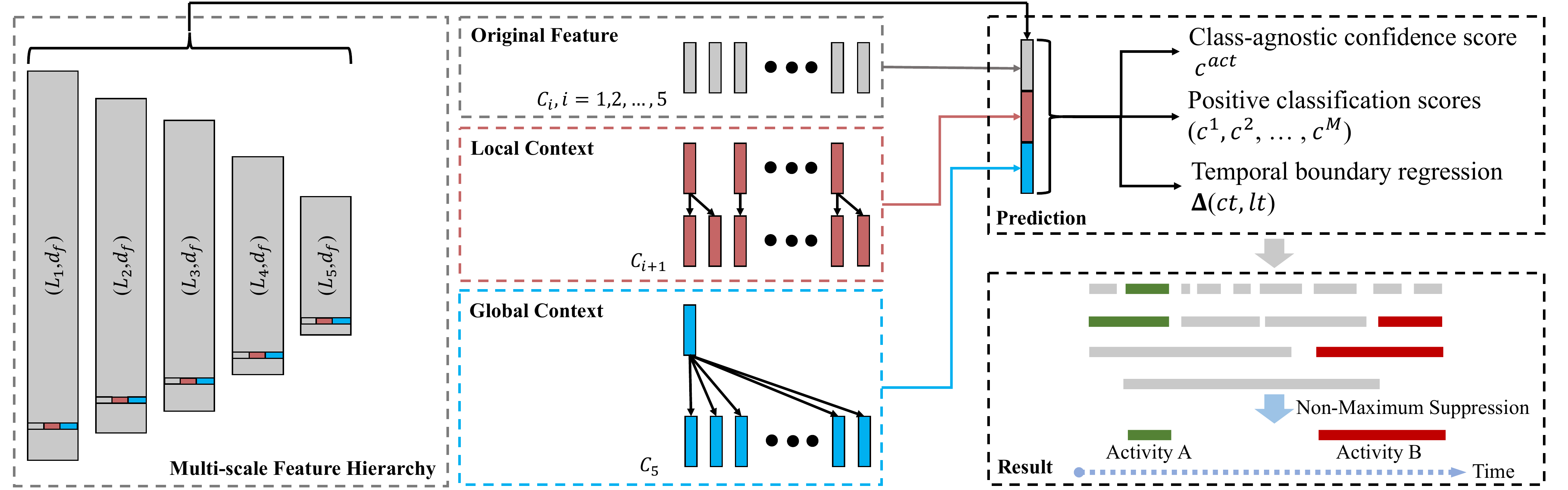}
\end{center}
   \caption{An illustration of local and global contexts when setting $N$ to $5$. Every temporal feature map cell at all scales is enhanced by its local context (next scale) and global context (last scale) to produce a set of prediction parameters. Temporal NMS is applied to produce the final detection results.}
\label{fig:context}
\end{figure*}
Our multi-scale feature hierarchy can easily incorporate contextual information since it naturally summarizes temporal information at different scales. To exploit the local temporal context for a specific layer $C_i$, we combine each temporal feature cell at $C_i$ with a corresponding feature cell at $C_{i+1}$. Specifically, we first duplicate each feature cell at $C_{i+1}$ twice to match with the temporal dimension of $C_i$ and concatenate the feature maps together. Thus, at each feature cell location in $C_i$, it not only contains feature at its original location but also a local context feature at the next scale. To exploit the global temporal context, instead of looking at the feature map in the next scale, we combine the feature with the last feature map $C_N$ which summarizes the whole video content. Similar to the local temporal context, we duplicate $C_N$ to have the same temporal dimension with $C_i$. We exploit local and global contexts at all layers in our network, thus, each temporal feature cell is enhanced by its local and global temporal information. We illustrate this mechanism in Fig.~\ref{fig:context}.

Each temporal feature map can produce a fixed set of detection predictions using a set of Conv1D layers. These are indicated on top of the feature network in Fig.~\ref{fig:context}. The basic operation for predicting parameters of a potential temporal detection is a Conv1D filter that produces scores for activity presence ($c^{act}$) and categories ($c^{1}$ to $c^{M}$, where $M$ is the total number of classes), and temporal offsets ($\Delta ct, \Delta lt$) relative to the default temporal location. The temporal detections at all scales are combined through temporal non-maximum suppression for generating the final detection results. 

\noindent \textbf{Comparison with previous works.} Neither  Zhang~\emph{et al.}~\cite{zhang2018s3d} nor Lin~\emph{et al.}~\cite{lin2017single} exploited any context features in their network. Dai~\emph{et al.}~\cite{dai2017temporal} included context features in the proposal stage, but they pooled features from different scales. Chao~\emph{et al.}~\cite{chao2018rethinking} only exploited the local temporal context. Our work considers both local and global temporal contexts and inherently extract contexts from a multi-scale temporal feature hierarchy. 

\section{Experiments}
We evaluate the proposed framework on the ActivityNet~\cite{caba2015activitynet} large-scale temporal activity detection benchmark. As shown in the experiments, our DTPN achieves state-of-the-art performance. We also perform a set of ablation studies to analyze the impact of different components in our network.

\subsection{Experimental Settings}
\noindent \textbf{Dataset.} ActivityNet~\cite{caba2015activitynet} is a recently released dataset which contains 200 different types of activities and a total of 849 hours of videos collected from YouTube. ActivityNet is the largest benchmark for temporal activity detection to date in terms of both the number of activity categories and number of videos, making the task particularly challenging. There are two versions, and we use the latest version $1.3$ which contains 19994 untrimmed videos in total and is divided into three disjoint subsets, training, validation, and testing by a ratio of $2:1:1$. On average, each activity category has $137$ untrimmed videos. Each video on average has $1.41$ activities which are annotated with temporal boundaries.  Since the ground-truth annotations of test videos are not public, following traditional evaluation practices on this dataset, we use the validation set for ablation studies.

\noindent \textbf{Evaluation Metrics.} ActivityNet dataset has its own convention of reporting performance metrics. We follow their conventions, reporting mean average precision (mAP) at different IoU thresholds $0.5$, $0.75$ and $0.95$. The average of mAP values with IoU thresholds $[0.5:0.05:0.95]$ is used to compare the performance between different methods.

\subsection{Implementation Details}
\textbf{Feature Extractor.} To extract the feature maps, we first train a Residual 3D ConvNet (Res3D) model~\cite{tran2018closer} on the Kinetics activity classification dataset~\cite{carreira2017quo}. The model takes as input a stack of 8 RGB frames with spatial size $256\times 256$, performs 3D convolutions, and extracts a feature vector with $d=2048$ as the output of an average pooling layer. We decode each video at 30 FPS to take enough temporal information into account, and each frame is resized to $256\times 256$. We set $K_1=L_1=16$ and $S=5$ for dynamic sampling, thus, we divide the input frame sequence into a set of $\{16, 32, 64, 128, 256\}$ segments and a snippet of window size $w=8$ is sampled in each segment. Each snippet is then fed into our Res3D model to extract a pyramidal input feature. Note that feature extraction can be done very efficiently with a single forward pass in batches.

\noindent \textbf{Temporal Anchors.} 
In our design, we associate a set of temporal anchors with each temporal feature map cell in the multi-scale feature hierarchy $\{C_i\}_{i=1}^{5}$. As described in Sec.~\ref{subsec:fusion}, the temporal dimension of $C_i$ is given as $L_i=2^{5-i},i\in [1,5]$. Regarding a feature map $C_i$, we set the length of each temporal anchor to be $\frac{1}{L_i}$ (as the input video length is normalized to $1$), and the centers are uniformly distributed with a temporal interval of $\frac{1}{L_i}$ in between. Thus, we assign a set of $\{16,8,4,2,1\}$ temporal anchors in our network which correspond to anchors of duration between $\frac{1}{16}$ and the whole video length. This allows us to detect activity instances with varying scales.

\noindent \textbf{Network Configurations.} 
Our system is implemented in TensorFlow and its source code will be made publicly available. All evaluation experiments are performed on a work station with NVIDIA GTX 1080 Ti GPUs. For multi-scale feature hierarchy, we generate a set of features with temporal dimension $\{16, 8, 4, 2, 1\}$ through both temporal convolution branch and temporal pooling branch as described in Sec.~\ref{subsec:fusion}.  In temporal convolution branch, we set the number of filters to $64$ for five different input features, and $d_t=320$ for all convolutional layers after concatenation. When training the network, we randomly flip the pyramidal input feature along temporal dimension to further augment the training data. The network is trained with multi-task end-to-end loss functions involving a regression loss, a classification loss and a localization loss. The whole network is trained for 20 epochs with the learning rate set to $10^{-4}$ for the first $12$ epochs and $10^{-5}$ for the last $8$ epochs. 

\subsection{Comparison with State-of-the-art}

Table~\ref{tb:ActNet} shows our activity detection results on the ActivityNet v1.3 validation subset along with state-of-the-art methods~\cite{singh2016untrimmed,wang2016uts,shou2017cdc,xu2017r,dai2017temporal,chao2018rethinking} published recently. The proposed framework, using a single model instead of an ensemble, is able to achieve an average mAP of $25.72$ that tops all other methods and perform well at high IoU thresholds, \emph{i.e.}, $0.75$ and $0.95$. This clearly demonstrates the superiority of our method. 

\setlength{\tabcolsep}{4pt}
\begin{table}
\begin{center}
\caption{Activity detection results on ActivityNet v1.3 validation subset. The performances are measured by mean average precision (mAP) for different IoU thresholds and the average mAP of IoU thresholds from $0.5:0.05:0.95$. }
\label{tb:ActNet}
\begin{tabular}{K{4.5cm}|K{1.3cm} K{1.3cm} K{1.3cm} K{1.3cm}}
\hline
IoU threshold & 0.5 & 0.75 & 0.95 & Average \\
\hline
Singh and Cuzzolin~\cite{singh2016untrimmed} (2016) & 34.47 & - & - & - \\
Wang and Tao~\cite{wang2016uts} (2016)  & 45.10 & 4.10 & 0.00 & 16.40 \\
Shou~\emph{et al.}~\cite{shou2017cdc} (2017) & 45.30 & 26.00 & 0.20 & 23.80 \\
\hline
Xu~\emph{et al.}~\cite{xu2017r} (2017) & 26.80 & - & - & 12.70 \\
Dai~\emph{et al.}~\cite{dai2017temporal} (2017) & 36.44 & 21.15 & \textbf{3.90} & - \\
Chao~\emph{et al.}~\cite{chao2018rethinking} (2018) & 38.23 & 18.30 & 1.30 & 20.22 \\
\hline \hline
DTPN (ours) & \textbf{41.44} & \textbf{25.49} & 3.26 & \textbf{25.72} \\
\hline
\end{tabular}
\end{center}
\end{table} 
\setlength{\tabcolsep}{1.4pt}

Note that the top half in Table~\ref{tb:ActNet} are top entries for challenge submission: our method is worse than~\cite{wang2016uts} at IoU threshold $0.5$ but their method is optimized for 0.5 overlap and its performance degrades significantly at high IoU thresholds, while our method achieves much better results ($25.49$ vs. $4.10$ at IoU threshold $0.75$); Shou~\emph{et al.}~\cite{shou2017cdc} builds a refinement network based on the result of~\cite{wang2016uts}, although they are able to improve the accuracy our method is still better when measured by the average mAP ($25.72$ vs. $23.80$). We believe the performance gain comes from our advanced temporal modeling design for both feature extraction and feature fusion, as well as rich temporal contextual information.

\subsection{Ablation Study}
\label{subsec:ablation}
To understand DTPN better, we evaluate our network with different variants on ActivityNet dataset to study their effects. For all experiments, we only change a certain part of our model and use the same evaluation settings. We compare the result of different variants using the mAP at $0.5$, $0.75$, $0.95$ and the average mAP. For a fair comparison, we don't concatenate contextual features in all experiments unless explicitly noted.

\setlength{\tabcolsep}{4pt}
\begin{table}
\begin{center}
\caption{Results for using a single-resolution feature map as the network input.}
\label{tb:single}
\begin{tabular}{K{4.5cm}|K{1.3cm} K{1.3cm} K{1.3cm} K{1.3cm}}
\hline
IoU threshold & 0.5 & 0.75 & 0.95 & Average \\
\hline
Single-256 & 36.75 & 22.09 & 1.94 & 22.18 \\
Single-128 & 36.93 & 21.93 & 2.86 & 22.32 \\
Single-64 & 35.47 & 21.39 & 2.56 & 21.63 \\
Single-32 & 35.62 & 21.78 & 2.57 & 21.66 \\
Single-16 & 33.64 & 20.69 & 1.82 & 20.63 \\
\hline
Pyramidal Input & \textbf{38.89} & \textbf{23.82} & \textbf{3.25} & \textbf{24.07} \\
\hline
\end{tabular}
\end{center}
\end{table} 
\setlength{\tabcolsep}{1.4pt}
\noindent \textbf{Dynamic Sampling vs. Single-resolution Sampling.} A major contribution of DTPN is using dynamic sampling to extract a pyramidal input feature as the network input. However, as a general SSD based temporal activity detector, single-resolution feature can also be applied as the input to our network. We validate the design for dynamic sampling pyramidal input by comparing with single-resolution sampling input: we keep the multi-scale feature network with $5$ temporal dimensions from $16$ to $1$ and the two-branch architecture, but instead of taking the pyramidal feature as input we only input a separate feature map of temporal size $256$, $128$, $64$, $32$ and $16$ independently. The hidden dimension for each layer is kept the same for a fair comparison. The results are reported in Table~\ref{tb:single}. Pyramidal input performs uniformly the best compared to single input, despite the network design, this clearly demonstrates the importance of multi-scale pyramidal feature extraction.

\setlength{\tabcolsep}{4pt}
\begin{table}
\begin{center}
\caption{Results for combing multiple feature maps as the network input.}
\label{tb:combine}
\begin{tabular}{K{1.3cm} K{1.3cm} K{1.3cm} K{1.3cm} K{1.3cm}|K{3.0cm}}
\hline
256 & 128 & 64 & 32 & 16 & Average mAP\\
\hline
\checkmark & \checkmark &  &  &  & 22.52 \\
  &  &  & \checkmark & \checkmark & 22.01 \\
\checkmark &  & \checkmark &  & \checkmark & 23.11 \\
\checkmark & \checkmark & \checkmark & \checkmark & \checkmark & \textbf{24.07} \\
\hline
\end{tabular}
\end{center}
\end{table} 
\setlength{\tabcolsep}{1.4pt}
\noindent \textbf{Multi-scale Feature Fusion.} We further validate our design to combine multiple features as our network input. Instead of just using a single-resolution feature as input, we investigate the effects of combining different input features. We also keep the same hidden dimension for each layer for a fair comparison. Table~\ref{tb:combine} compares different combination schemes: we observe that only dense sampling ($256$+$128$) or sparse sampling ($32$+$16$) leads to inferior performance compared to sampling both densely and sparsely ($256$+$64$+$16$); By adding more fine-grained details ($128$ and $32$), our pyramidal input achieves the best result.

\setlength{\tabcolsep}{4pt}
\begin{table}
\begin{center}
\caption{Results for the impact of the two-branch network architecture.}
\label{tb:branch}
\begin{tabular}{K{4.5cm}|K{1.3cm} K{1.3cm} K{1.3cm} K{1.3cm}}
\hline
IoU threshold & 0.5 & 0.75 & 0.95 & Average \\
\hline
TConv & 27.12 & 14.70 & 1.34 & 15.12 \\
TPool & 29.77 & 17.24 & 2.16 & 17.12 \\
\hline
TConv+TPool (two-branch) & \textbf{38.89} & \textbf{23.82} & \textbf{3.25} & \textbf{24.07} \\
\hline
\end{tabular}
\end{center}
\end{table} 
\setlength{\tabcolsep}{1.4pt}
\noindent \textbf{Two-branch vs. Single-branch.} Here, we evaluate the impact of the two-branch network architecture. In our design, We propose to use a separate temporal convolution branch and temporal pooling branch and fuse the two feature hierarchies at the end. However, either branch can be used independently to predict the final detection results. Table~\ref{tb:branch} lists the performance of models with temporal convolution branch only (TConv) and temporal pooling branch only (TPool). We conclude that two-branch architecture can significantly improve the detection performance (more than $5\%$ in comparison with single-branch). 

\setlength{\tabcolsep}{4pt}
\begin{table}
\begin{center}
\caption{Results for incorporating local and global temporal contexts.}
\label{tb:context}
\begin{tabular}{K{4.5cm}|K{1.3cm} K{1.3cm} K{1.3cm} K{1.3cm}}
\hline
IoU threshold & 0.5 & 0.75 & 0.95 & Average \\
\hline
w/o Context & 38.89 & 23.82 & 3.25 & 24.07 \\
w/ Local Context & 40.01 & 24.50 & 3.24 & 24.70 \\
w/ Global Context & 40.17 & 24.20 & \textbf{3.54} & 24.62 \\
\hline
w/ Local+Global Contexts & \textbf{41.44} & \textbf{25.49} & 3.26 & \textbf{25.72} \\
\hline
\end{tabular}
\end{center}
\end{table} 
\setlength{\tabcolsep}{1.4pt}
\noindent \textbf{Local and Global Temporal Contexts.} We contend that temporal contexts both locally and globally are crucial for temporal activity detection. Since local and global contextual features are extracted from different layers and combined through concatenation, we can easily separate each component and see its effect. As reported in Table~\ref{tb:context}, We compare four different models: (1) model without temporal context (w/o Context); (2) model only incorporating local context (w/ Local Context); (3) model only incorporating global context (w/ Global Context); (4) model incorporating both local and global contexts (w/ Local+Global Contexts). We achieve higher mAP nearly at all IoU thresholds when incorporating either local or global context, and we can further boost the performance by combining both contexts at the same time.

\begin{figure*}[t!]
\begin{center}
\includegraphics[width=1.0\linewidth]{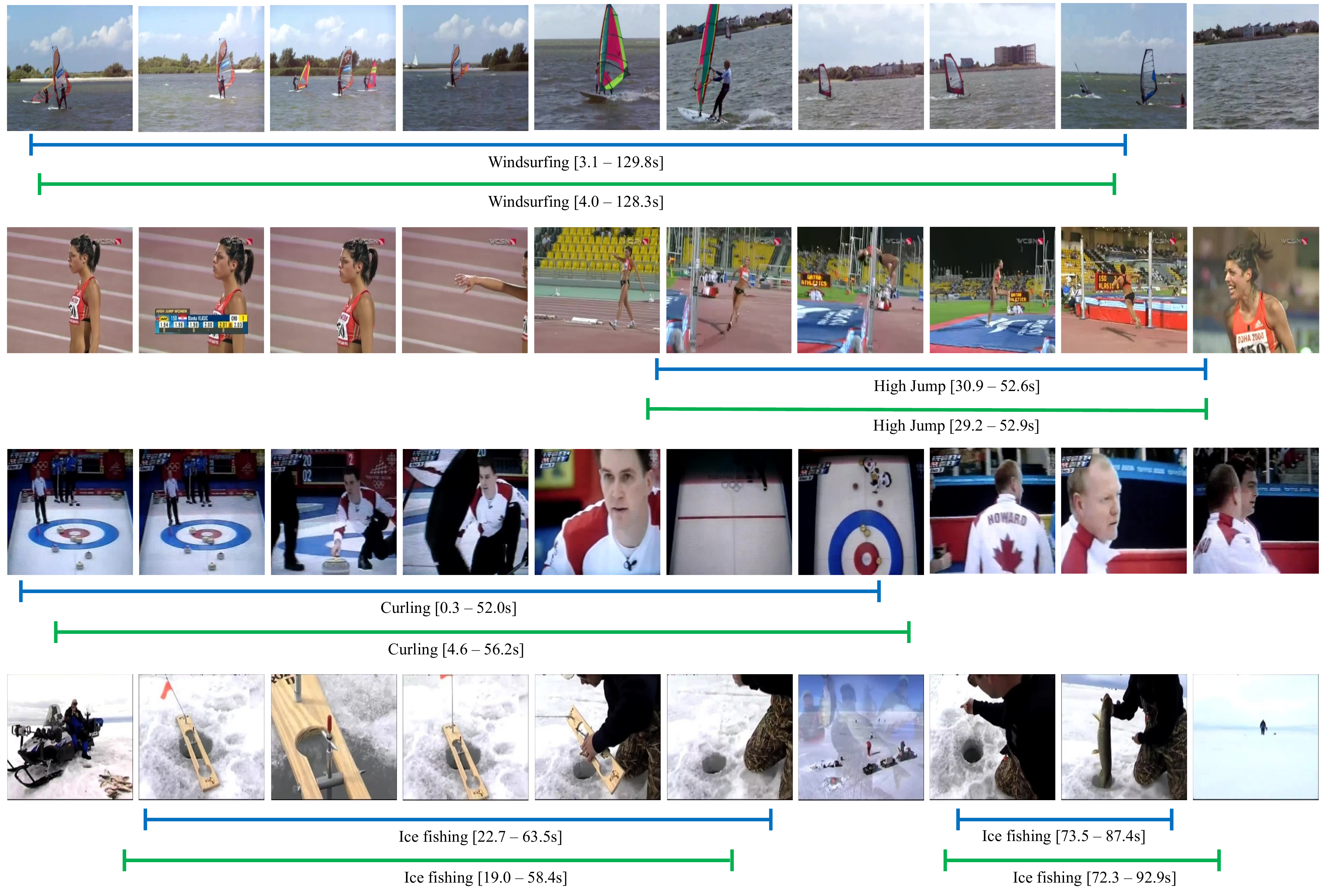}
\end{center}
   \caption{Qualitative visualization of the top detected activities on ActivityNet. Each sequence consists of the ground-truth (blue) and predicted (green) activity segments and class labels.}
\label{fig:qualitative}
\end{figure*}

\noindent \textbf{Qualitative Results.} We provide qualitative detection results on ActivityNet to demonstrate the effectiveness and robustness of our proposed DTPN. As shown in Fig.~\ref{fig:qualitative}, different video streams contain very diversified background context and different activity instances vary a lot in temporal location and scale. DTPN is able to predict the accurate temporal span as well as the correct activity category, and it is also robust to detect multiple instances with various length in a single video.

\begin{table}
\begin{center}
\caption{Comparison of our approach and the state-of-the-art methods in the approximate
computation time(s) to process each video on ActivityNet dataset.}
\label{tb:time}
\begin{tabular}{K{1.8cm}|K{2.5cm} K{2.0cm} K{3.0cm} K{2.0cm}}
\hline
Method & Shou~\emph{et al.}~\cite{shou2017cdc} & Xu~\emph{et al.}~\cite{xu2017r} & Mahasseni~\emph{et al.}~\cite{mahasseni2017budget} & DTPN(ours) \\
\hline
Time (s) & $> 930$ & $3.2$ & 0.35 & \textbf{0.5} \\
\hline
\end{tabular}
\end{center}
\end{table} 

\noindent \textbf{Activity Detection Speed.} We benchmark our network on a single GTX 1080 Ti GPU to measure the activity detection speed. One activity detection in our framework is measured as a single forward-pass of the whole network, and we follow the same strategy reported in~\cite{mahasseni2017budget} to calculate the approximate detection time for different methods. In Table~\ref{tb:time}, we compare our approach with the state-of-the art methods in the approximate computation time to process each video. Due to the single-shot end-to-end design with simple Conv3D building blocks, our DTPN is very efficient and can process a single video in $0.5s$ which is significantly faster than most state-of-the-art methods~\cite{xu2017r,shou2017cdc}. 

\section{Conclusions}
In this paper, we introduce DTPN, a novel network architecture specifically designed to address three key challenges arising from the scale variation problem for temporal activity detection. DTPN employs a multi-scale pyramidal structure with three novel architectural designs: 1) pyramidal input feature extraction with dynamic sampling; (2) multi-scale feature hierarchy with two-branch network; and (3) local and global temporal contexts. We achieve state-of-the-art performance on the challenging ActivityNet dataset, while maintaining an efficient single-shot, end-to-end design.


\bibliographystyle{splncs04}
\bibliography{0034}

\end{document}